\documentclass{article}
\usepackage[pdftex]{graphicx}
\usepackage{authblk}
\usepackage{my_symbols}
\usepackage{subcaption}
\usepackage{url}
\usepackage{color}			

\begin{document}

\title{Probabilistic Regressor Chains\\ with Monte Carlo Methods}
\author[1]{Jesse Read}
\author[2]{Luca Martino}
\affil[1]{LIX, Ecole Polytechnique, Institut Polytechnique de Paris, France}
\affil[2]{Dept.\ of Signal Theory and Comm., Univ.\ Carlos III de Madrid}

\date{}
\maketitle


\begin{abstract}
	A large number and diversity of techniques have been offered in the literature in recent years for solving multi-label classification tasks, including classifier chains where predictions are cascaded to other models as additional features. The idea of extending this chaining methodology to multi-output regression has already been suggested and trialed: regressor chains. However, this has so-far been limited to greedy inference and has provided relatively poor results compared to individual models, and of limited applicability. 
In this paper we identify and discuss the main limitations, including an analysis of different base models, loss functions, explainability, and other desiderata of real-world applications. To overcome the identified limitations we study and develop methods for regressor chains. In particular we present a sequential Monte Carlo scheme in the framework of a probabilistic regressor chain, and we show it can be effective, flexible and useful in several types of data. We place regressor chains in context in general terms of multi-output learning with continuous outputs, and in doing this shed additional light on classifier chains. 
\end{abstract}



\section{Introduction}
\label{sec:intro}

Multi-dimensional data is ever-more present in industrial and scientific contexts. For example, multi-label classification has made a significant impact in the machine learning literature over recent years, where data points are naturally associated with multiple outputs\footnote{Distinguished from multi-class classification in that a single output may take multiple values -- but only one such value is assigned per output}. Rather than a naive method of building one model per output; advanced methods can model the outputs together, resulting in better predictive performance. Although the potential of individual models is periodically revived under particular scenarios, the vast majority of literature proposes joint modeling, and as a result show improvement in both predictive performance and efficiency. A recent review of this area containing many useful references, is given in \cite{UnifiedView}.  

\begin{figure}
	\centering
	\includegraphics[scale=1]{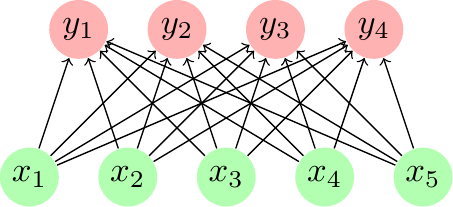}
	\includegraphics[scale=1]{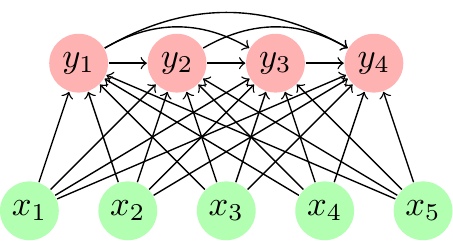}
	\caption{\label{fig:cc}The naive vs chaining models. Each target $y_j$ is learned by a base model, where inputs are shown as incoming arrows, and output/prediction as an outgoing arrow.}
\end{figure}

An established method in multi-label classification is that of \textit{classifier chains}, where a model is trained for each label, but estimates of the other models are used as additional features, in a cascaded chain along the target labels. This is exemplified (and contrasted to the naive approach) in \Fig{fig:cc}. 

This `chaining' mechanism, although simple in its basic form, has proved successful in multi-label classification and provided dozens of modifications, extensions, and improvements, in the literature (see, e.g., \cite{ECC2,PCC,MCC2,OnLabelDependence2,UnifiedView, READ20152096,READ201745} and references therein).

It is flexible, in the sense that any base model case be used to predict each label, and powerful, in the sense that even relatively simple base models lead to greater predictive performance than if they had been used separately. 


Given its impact in multi-label classification, the motivation of this paper is to more-thoroughly investigate the use of the chaining approach in the regression context with continuous output variables. This investigation is needed since only greedy inference can be applied directly, but even in this case the application of chaining to the regression context is not straightforward and the effectiveness of such application is not yet widely explored. 

After summarizing background work (\Sec{sec:background}), we provide a rigorous discussion and development of probabilistic regressor chains, including a survey of possible approaches (Sections \ref{sec:RC} and \ref{sec:PRC}). Following this, we develop a sequential Monte Carlo scheme (\Sec{sec:smcrc}). This scheme allows sampling of candidate paths through the target space, which is useful for many applications. We implement and test several of these on synthetic and real-world datasets involving multiple continuous outputs (\Sec{sec:experiments}), the results of which we reflect upon in detailed discussion (\Sec{sec:discussion}). After looking at connections to related and potential future work (\Sec{sec:related}) we draw conclusions and make recommendations (\Sec{sec:conclusion}). 



\section{Chain Methods}
\label{sec:background}

Given a dataset $\D = \{(\x_i,\y_i)\}_{i=1}^N$, where $\x_i \in \mathcal{X} \subseteq \mathbb{R}^D$ and $\y_i \in \Y \subseteq \mathbb{R}^L$, we are interested building a model that can provide predictions corresponding to $L$ target variables\footnote{In cases where it is not clear from context, we will denote $y_{ij}$ as the value of the $j$-th attribute of the $i$-th instance}, $\ypred = [\yp_1,\ldots,\yp_L]$, for any given input observation $\x = [x_1,\ldots,x_D]$. 

In the multi-output \emph{classification} scenario \cite{ECC2,PCC,MCC2,CCAnalysis}, the $j$-th output takes some value $y_j \in \{1,\ldots,K_j\}$ (we may highlight the popular case of binary outputs where each $y_i \in \{0,1\}$, known often as \emph{multi-label classification}) or -- in the case of continuous outputs, i.e., multi-output \emph{regression} -- then each $y_j \in \R$. 


A naive approach is to simply build a separate model for each target independently, such that
\begin{equation}
	\label{eq:BR}
	\ypred = [\yp_1,\ldots,\yp_L] = [h_1(\x),\ldots,h_L(\x)]
\end{equation}
where each model $h_j$ has been built from training set $\{(\x_i,y_{ij})\}_{i=1}^N$, as a traditional single-output classifier or regressor according to the domain of $Y_j$ (clearly, a regressor, if $Y_j \in \R$). The exact class of model (type of regressor) is largely dependent on preference and/or driven by domain assumptions and constraints.   


Particularly with regard to multi-output classification, a large volume of literature proposes a plethora of models that out-compete this baseline by modelling the dependencies among outputs; consider \cite{ECC2,PCC,RAkEL2,MCC2,OnLabelDependence2,CCAnalysis,UnifiedView} and references therein. The multi-output regression literature is smaller in volume (consider, e.g., \cite{HanenMORSurvey,MTRexpansion}), being both more recently considered for specialized methods, and being more difficult to obtain better results with such methods -- precisely a motivating factor behind this work. 

%

%
%

\label{sec:chaining}


In prior work (\cite{ECC2,PCC,MCC2,CCAnalysis}, and many others), a chaining mechanism was studied for multi-label classification, known as \textit{classifier chains}. Due to the relatively large amount of work in the literature on this topic, including continued recent interest (e.g., \cite{CECC,CCnet} among many others), we argue it is worth studying this mechanism in particular with application to multi-output regression, which we will denote \textit{regressor chains}. 

It is useful to study regressor chains in light of previous developments in classifier chains. In its simplest form, the estimates of other labels are used as feature inputs for the following classifier, thus augmenting the input space along the chain. Given some test instance $\x$, we may obtain an estimate of $\ypred = [\yp_1,\ldots,\yp_L]$ as
\begin{equation}
	\label{eq:cc}
	\ypred = [h_1(\x), h_2(\x,\yp_1), \ldots, h_L(\x,\yp_1,\ldots,\yp_{L-1})]
\end{equation}
where a recursion takes placed based on $\yp_j = h_j(\ldots)$. 

In the classification context, this is known as a \textit{classifier chain} and, specifically, one with \textit{greedy inference}. Note that models $h_j$ can be estimated individually and in parallel at training time, similarly to those in \Eq{eq:BR}, and -- likewise -- those models may be any class appropriate for each individual target (a binary classifier for binary output, etc). 

In this case of greedy inference, a single path/combination is considered. However, note that this $\ypred$ in particular is just one possible prediction, and not necessarily the best. The addition of probabilistic inference was an important development \cite{PCC,OnLabelDependence2}, where a search is carried out, typically to obtain a MAP estimate\footnote{Although configurations for other losses are also possible; see \cite{PCC}}, thus maximizing $0/1$ loss; hence
\begin{align}
	\ypred = \h(\x) &= \argmax_{\y \in \Y} P(\y|\x) \notag \\
					&= \argmax_{\y \in \Y} \prod_{j=1}^L P(y_j|\x,y_1,\ldots,y_{j-1}), \label{eq:pcc}
\end{align}
where $\Y$ is the space of possible paths over the $L$ outputs and $P(y_j|\x,\cdot)$ is a pmf associated with the $j$-th model (henceforth we use the abbreviation $P_j$ where suitable). This is known as a \textit{probabilistic classifier chain}. The need for a probabilistic interpretation (i.e., pmfs $P_j$) can be addressed by building such models; logistic regression is a common choice. 

One may observe that the search space in \Eq{eq:pcc} is exponential with the number of labels $L$. Indeed, complete inference (all branches of the probability tree are explored) is usually prohibitive. Therefore, one may consider the search space as a probability tree, and conduct a tree search, where each $P_j$ provides a weight on each branch. \Fig{fig:eg} offers some visual intuition. 

\begin{figure}[h]
	\centering
	\includegraphics[width=0.3\textwidth]{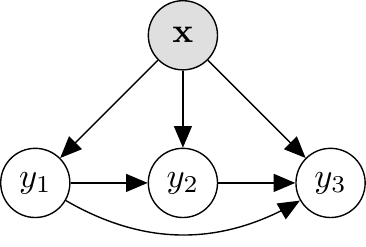}
	\hspace{1cm}
	\includegraphics[width=0.3\textwidth]{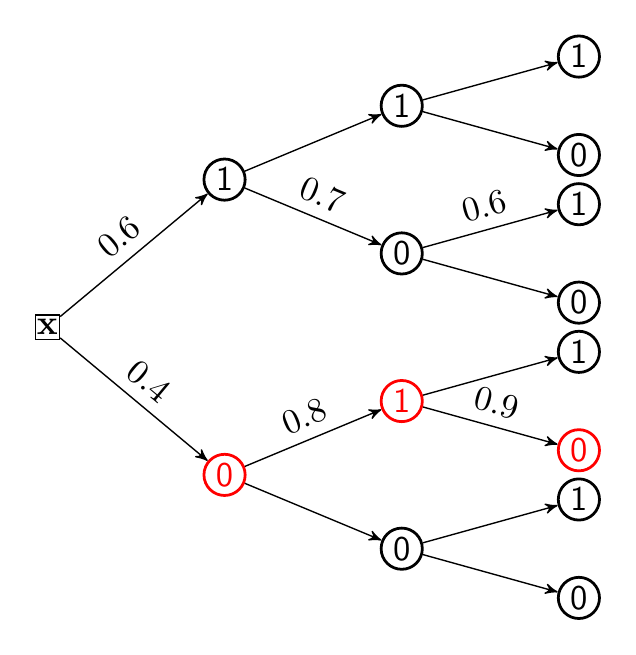}
	\caption{\label{fig:eg}Probabilistic classifier chains where $L=3, y_j \in \{0,1\}$: As a probabilistic graphical model (left), and with two explored paths in $\Y^L$ in the probability tree. Note that the best path (in {\color{red}red}, right) is not found by greedy inference.  There are $2^L$ possible paths ($\Y = \{0,1\}^3$). The label on each edge indicates $P_j(y_j=1)$ (shown for explored paths).}
\end{figure}

Many search methods have been applied for this purpose (a survey is given in \cite{AstarCC}). In particular we highlight the approach of Monte-Carlo sampling \cite{MCC2} as -- unlike many of the search approaches -- is directly applicable to regressor chains, and as such is most relevant to the material developed in \Sec{sec:smcrc} of this work. In this approach, $M$ samples are taken, $m=1,\ldots,M$, and weighted as 
\begin{align}
	y^{(m)}_j     & \sim P(y_j|\x,y^{(m)}_1,\ldots,y^{(m)}_{j-1}) \label{eq:f_dyn_luca} \\
		w^{(m)}_j & = w^{(m)}_{j-1} \cdot P(y^{(m)}_j|\x,y^{(m)}_1,\ldots,y^{(m)}_{j-1}) \label{eq:f_obs_luca} 
\end{align}
where $w^{(m)}_0 = 1$. An example is shown in \Fig{fig:eg} where $M=2$. A final prediction is obtained as 
\begin{align}
	\ypred      & = \y^{(m^*)}  = \argmax_{\y \in \{\y^{(m)}\}_{m=1}^M}  P(\y|\x)  \label{eq:PFE_luca_correct}
\end{align}
where $m^* = \argmax_{m \in \{1,\ldots,M\}} w_L^{(m)}$ (index of the maximum weight), and where $\y^{(m)} = [y^{(m)}_1,\ldots,y^{(m)}_{L}]$ is a complete sequence across label indices. Note that complexity is determined by $M \ll 2^L$. 



It has been noticed that a chaining procedure can be applied in an off-the-shelf manner to the multi-output regression context involving continuous targets, namely with greedy inference; see, e.g., \cite{HanenMORSurvey}. However, as we highlight and discuss in the following section, even though the application is identical, there are some major differences in the way regressor chains behave and the relative results they obtain (with respect to non-chained methods). 



\section{The Poor Behaviour of Regressor Chains}
\label{sec:RC}


Applying greedy inference in chains in the case of regression is -- exactly as in classification --  a case of each output simply being ``plugged in'' to the following model as an additional feature. Recall that this simply means that predictions $\yp_1,\ldots,\yp_{j-1}$ are treated as fixed observations (i.e., and not random variables) when inferring $y_j$. Several papers have trialed this approach, e.g., \cite{HanenMORSurvey}. Unfortunately, predictive performance is not as impressive wrt independent regression models. We explain why. 

Recall \Eq{eq:cc} from above. A first choice of model in the regression context may be least squares where $\yp_j = h_j(\x) = \w_j^\top[x_1,\ldots,x_D,\yp_1,\ldots,\yp_{j-1}]$. Plugging this in to the equation, and supposing for simplicity $D=1$ input and $L=2$ outputs, we observe that\footnote{Let $\w_{j} = [w_{j,1},w_{j,2}]$}

\begin{align*}
	\yp_2 &= h_2(x,\yp_1) \\
		  &= \w_2^\top[x,\yp_j]  \\
		  &= \w_2^\top[x,w_1 x]  \\
		  &= w_{2,1}x + w_{2,2} w_1 x  \\
		  &= x( w_{2,1} + w_{2,1} w_1)  = w' x 
\end{align*}
and thus clearly we see that label predictions $\yp_1,\ldots,\yp_{j-1}$ are superfluous in predicting $\yp_j$; the regressor chain in this context simply performs a series of linear transformations which can naturally be represented as a single transformation. 

\begin{figure}[h]
	\centering
	\includegraphics[scale=1]{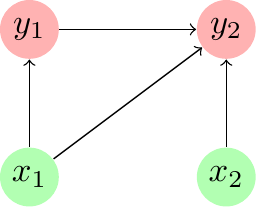}
	\caption{\label{fig:nonisotropic} Even when $x_2$ arrives only at timestep $2$, information can still be carried forward from $x_1$ (rather than via $y_1$), thus making the label cascade superfluous wrt the prediction $\yp_2$ as long as $f(\x)$ is well modeled, even in this case.}
\end{figure}

It may be tempting to argue for application to the case where $x_j$ is observed only at \emph{time} $x_j$. In this scenario we presume that inputs are received over time, thus estimating $\yp_j = h(x_1,\ldots,x_j,\yp_1,\ldots,\yp_{j-1})$. Unfortunately, we may show that $\yp_1,\ldots,\yp_{j-1}$ are still not needed wrt $\yp_j$ since all available information already comes from $x_1,\ldots,x_j$ directly via earlier labels. This is seen clearly in illustration; \Fig{fig:nonisotropic}. 

Another fundamental issue is the selection of loss metric. An obvious and popular choice for regression is based on the mean squared error (MSE) loss criterion (as indeed considered in the earlier experimentation of regressor chains in \cite{HanenMORSurvey,MTRexpansion} among others). 


Under random variables $\rY = Y_1,\ldots,Y_L$, we see that the minimum MSE (MMSE) estimator under observation $\x$ is given as follows, 

\begin{eqnarray}
	\vmuest &=& \Exp[Y_1,Y_2,\ldots,Y_L| \x]  \notag \\ 
	&=& \Exp[\rY|\x]=\int \y p(\y|\x) d\y \label{eq:expectation} \\
						&=& \int \y \frac{p(\y,\x)}{p(\x)} d\y  \notag \\ 
						&=& \int \y \frac{p(y_1|\x) \prod_{j=2}^L p(y_j|\x,y_1,\ldots,y_{j-1})p(\x)}{p(\x)} d\y  \notag \\
						&\propto& \int \y \cdot p(y_1|\x) \prod_{j=2}^L p(y_j|\x,y_1,\ldots,y_{j-1}) d\y. \notag  
						\end{eqnarray}
wrt marginal densities $p_j$ (homologous to the pmfs of \Eq{eq:joint}). 
Let us emphasise that $\vmuest = [\muest_1,\ldots,\muest_L]$ is a vector of $L$ integrals where
\begin{align}
	\muest_j            & =\int y_j p(\y|\x) d\y  \notag \\
	                 & =\int y_j p(y_1,\ldots,y_L|\x) dy_1 \cdots dy_L=\int y_j p_{j}(y_j|\x) dy_j \label{eq:si_number}
\end{align}

If we assume a Gaussian distribution $\epsilon_j \in \N(0,\sigma_j^2)$ (a natural view of ordinary least squares \cite{EoSL}) then again we may connect to the case discussed at the beginning of this section, where $\muest_1 = \w^\top_1\x$ and so on for $\muest_1,\ldots,\muest_L$. 

More generally, we are not only interested in obtaining point-wise estimators $\mu_j$, but we are also interested in extracting all the statistical information encoded within this posterior $p(\y|\x)$, such as uncertainty measures, credible intervals, and quantiles as well. So, our goal is to approximate complex integrals involving this posterior
\begin{equation}
	\label{eq:joint}
	p(\y|\x) = p(y_1,y_2,\ldots,y_{L}|{\x}) = p(y_1|\x) \prod_{j=2}^L p(y_j|\x,y_1,\ldots,y_{j-1})
\end{equation}

Having an estimate of the shape of $p$ then allows us to estimate other values aside from the expected value (i.e., \Eq{eq:si_number}) such as the median or the mode. 
We remark that classifier chains typically predicts a mode in the form of a MAP estimate.   


Notice, however, that -- in spite of using the chain rule to factorize the joint in this manner 
this estimate (and hence related integrals, \Eq{eq:si_number}) are generally intractable. This is unlike in probabilistic \emph{classifier} chains where each pmf $P_j$ models a variable taking a finite number of discrete values -- inherently more tractable in most cases.

In all cases an exhaustive search in the joint is intractable for a large number of outputs $L$, hence efficient tree search methods such as Monte Carlo sampling (again, refer to \Fig{fig:eg}). However, a tree cannot be formed on continuous output space, and hence no such search is initially possible in the regression context. 
Discussion of the output space also brings us to identify a further pitfall in regression chains: the possibility of extensive ``error propagation''. In classifier chains, this is well known, and easily detectable (e.g., in \label{fig:eg}). However, the $P_j \in [0,1]$ space  is inherently limited. On the other hand, under regressior chains, the estimate of greedy inference may completely degenerate and become lost in $p_j \in \R$ space, we progress down the chain towards $j=L$.

Therefore we have isolated and discussed some main drawbacks to the application of greedy regressor chains: the type of loss functions often considered in regression (e.g., MSE) totally change the behaviour of the chaining methodology, and with a linear regression scheme, such as least squares, there is no benefit to the chaining mechanism. Even worse, error propagation may be catastrophic.  

Having identified the lacking in regressor chains, as compared to their classification counterpart, in the following section we study the idea of probabilistic regressor chains which we later build on to suggest new methods. 


\section{Probabilistic Regressor Chains with Monte Carlo Search}
\label{sec:PRC}

In probabilistic classifier chains, inference may be cast as a search in the probability tree, as discussed above. Many tree-search methods are applicable. We gave an example of Monte Carlo search which takes $M$ weighted samples across the tree (recall \Eq{eq:f_dyn_luca} and \Eq{eq:f_obs_luca}). In the regression context there is no inherent tree to search. We could however create a tree by using particular values $y_j^{(m)} \in \R$ as nodes. This requires modeling of the marginals in such a way that we may draw samples 
	\begin{equation}
		\label{eq:luca_cazzo}
		y^{(m)}_j \sim p(y_j|\x,y_1^{(m)},\ldots,y_{j-1}^{(m)}),
	\end{equation}
and in particular these samples should be at an `interesting' part of the space, such as near the modes. 

\begin{figure}
	\centering
	\includegraphics[width=0.32\textwidth]{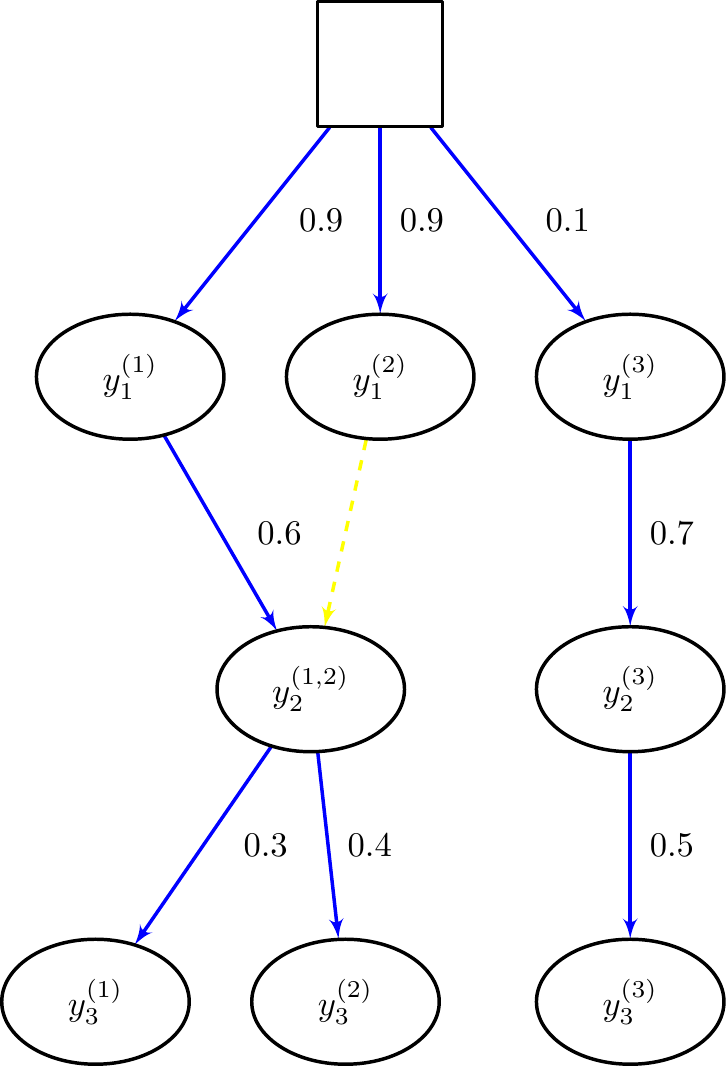}
	\caption{\label{fig:demo2b}A hypothetical search tree through particle space along the paths for $M=3$, $L=3$. Nodes show samples $y_j^{(m)}$ and branches show evaluation of pdf $p_j(y_j^{(m)}|\x,y_1^{(m)},\ldots,y_{j-1}^{(m)})$. The yellow branch indicates resampling (see \Sec{sec:smcrc}) however this is shown for illustration; this extra branch is not required for inference.}
\end{figure}

\Fig{fig:demo2b} shows an example illustration of samples $\{y_j^{(m)}\}$ forming paths and thus a tree with each complete path from root to leave as a possible prediction $\y \in \R^L$. 

This is an interesting idea, as with a cloud of samples, we can not only obtain a tree and apply existing methods used in the classification context,, but also easily obtain an approximation a MAP estimator as well, as in \Eq{eq:PFE_luca_correct}. Explicitly we can write this as:
\begin{equation}
	\label{eq:the_argmax}
	\y^{(m^*)}=\argmax \left\{p(\y^{(1)}|\x),\cdots,p(\y^{(M)}|\x)\right\},
\end{equation}
and more generally (estimators other than MAP), a prediction may be given as path 
\begin{equation}
	\label{eq:the_general}
	\ypred = g\left(\{\y^{(m)},\w^{(m)}\}_{m=1}^M\right) 
\end{equation}
chosen by function $g$ over a tree with branches defined across nodes $\{\y^{(m)}\}$ and weighted by $\{w^{(m)}_j\}$, i.e., holding the marginal branch cost from depth $j-1$ to $j$ by the $m$-th sample. Recall \Fig{fig:demo2b}. It means that \Eq{eq:the_argmax} makes use of the fact that $p(\y^{(m)}|\x) = \prod_{j=1}^L w^{(m)}_j$. In the remainder of this work we occasionally omit the subscript and use $w^{(m)} = p(\y^{(m)}|\x)$ as the path cost from root to a leaf, i.e., $w^{(m)} = p(\y^{(m)}|\x)$ where $p$ as in \Eq{eq:joint} (for notational simplicity).


Suppose we define a cost function $c$ based on path costs $w^{(m)}_j$. For example, cost is given whenever crossing a low-density region according to the estimated $p_j$s. 
To minimize expected loss, 
\begin{align*}
	\Exp_{\mY \sim \D} [c(\mY, \h(\x)) | \x] &= \int_{\dY} c(\y,\h(\x)) p(\y|\x) \d \y
\end{align*}
we use our estimation of $p$ (and decomposition as \label{eq:joint}) and Monte Carlo samples to replace the integral, thus giving something like \Eq{eq:the_argmax} (depending on the choice of loss function). 
	

A fundamental consideration is how to estimate each component $p_j$ of $p$, and how to draw samples from it. The major issue at stake is that that a powerful non-linear predictor is required to justify chains (as discussed above) yet such a predict may not be ideal for sampling from. We address this issue with a sequential Monte Carlo method, presented in the following section. But let us first look briefly at some available alternatives, all of which could be considered a variation of probabilistic regressor chains. 

\subsection{Discretization and classification}
\label{ssec:1}

A first approach to the inference problem is to simply discretize the label space and proceed from a classification perspective. This is not the same as taking samples (which is done at prediction time). As explained above, the classification perspective of chains is easier to justify, and a wealth of methods available. Of course not all problems are suited to discretization of the label space, but often it is suitable and even common practice, for example in the domain of reinforcement learning (see, e.g., \cite{MCTS_CAS}). 

In this case we are learning $p(y_j \in \dS^{(k)}_j | \z) \Leftrightarrow P(y_j = k)$ (the RHS relates to \Eq{eq:pcc}) where bin $\dS^{(k)}_j \subset \R$; and thus removing the need to estimate any integrals. Clearly it is fundamental to choose a suitable set of bins (which correspond to a finite set of class labels). 

\subsection{Bayesian regression}
From the perspective of a greedy chain (earlier labels are fixed input), we may elabourate $p_j$ as Bayesian linear regression:
\begin{equation}
	\label{eq:luca_cazzorola}
	p_j = 
	\N(y_j | \vmu_j, \mSigma_j)
\end{equation}
with sufficient statistics $\vmu_j$ and $\mSigma_j$. As a Gaussian, sampling from this distribution (e.g., as in \Eq{eq:luca_cazzo}) is straightforward. However, it based on a linear combination of $\vmu_j = \vtheta^\top\x$ and provides a unimodal Gaussian-shaped estimate, not suitable for many types of data.   

%


\subsection{Variational inference}

We may approximate each\footnote{As a reminder, $p_j = p(y_j|\x,y_1,\ldots,y_j)$} $p_j$ with some other distribution 
\(
	q(y_j | \theta)  \approx p_j 
\) 
as in variational Bayesian methods, which turns inference into an optimization problem
\[
	\theta^* = \argmin_{\theta} \textsf{KL}(q(y_j|\theta)\|p_j) 
\]
(minimizing Kullback-Leibler divergence); see, e.g., \cite{Barber}. We could then sample $y_j \sim q(y_j|\theta)$ instead. Unlike MCMC methods, this approach does not provide an exact model of $p_j$ in the limit (with sufficient samples). 

\subsection{Density estimation}
\label{sec:DE}

Let us denote $p_j = p(y_j|\x,y_1,\ldots,y_{j-1}) = p(y_j|\z)$. We may also model the target density $p(y_j|\z) = p(y_j,\z)/p(\z)$ 
using a non-parametric method such as a Parzen window (i.e., a kernel density estimate, KDE), where
\[
	p(y_j,\z) = \frac{1}{N} \sum_{i=1}^N K_h(\z,\z_i) K_h(y_j,y_i)\text{\quad and\quad}
	p(\z) = \frac{1}{N} \sum_{i=1}^N K_h(\z,\z_i) 
\]
for some kernel $K_h$ (of bandwidth parameter $h$). 

Sampling can be carried out for certain kernels such the Gaussian kernel. Although of course, the general disadvantages of kernel methods apply (quadratic complexity wrt number of examples, and difficulty in incremental modeling).


\section{Sequential Monte Carlo Regressor Chains}
\label{sec:smcrc}

The methods just described above are adequate to build a probabilistic regressor chain if the resulting approximation of $p_j$ is adequate for both sampling (obtaining $y^{(m)}_j$) and accurate evaluation (obtaining $w^{(m)}_j$). This restricts applicability considerably, and our selection is reduced especially with high-dimensional input observations. For example, many methods may provide an evaluation of a pdf, but not useful sampling. 

In this section we build a state-space model for probabilistic regressor chains, separating the sampling and evaluation functions. It is essentially a \emph{particle filter} (PF, see, e.g., \cite{Djuric03}). We make some particular considerations and adaptations for its application in a probabilistic regressor chain. 

\subsection{The particle filter}


To briefly review: a particle filter in a state space model consists of a model 
\begin{equation}
	\label{eq:model}
	\mathcal{M}:\left\{
		\begin{array}{l}
		 f(y_j|y_{j-1})\\
		 \ell(\x_j|y_{j})
		\end{array}
		\right.
\end{equation}
running over time-steps $j=1,\ldots,L$, encompassing the transition function and observation function, $f$ and $\ell$, respectively\footnote{Regarding notation, it is fundamental for those familiar with the literature on particle filters and continuous state-space models, to observe that in this work $y_j$ is the state, and $\x_j$ is the observation; as in line with the machine learning literature}. See \cite{Djuric03} for an in-depth introduction and survey. 

Under this notation, the vanilla particle filtering method for obtaining a marginal estimation (i.e., the prediction\footnote{In machine learning, prediction often refers to an \emph{estimation} for the current instance, rather than for some future time instance, as often in dynamical systems terminology}) for $y_j$ can be written as:
\begin{align}
	y^{(m)}_j          & \sim f(y| y_{j-1}) \label{eq:f_dyn} \\
		w^{(m)}_j      & = w^{(m)}_{j-1} \cdot \ell(y^{(m)}_j|\x) \label{eq:f_obs} \\
	\yp_j = \Exp[Y_j]          & \approx \frac{1}{M} \sum^M_{m=1}w^{(m)}_j y^{(m)}_j \label{eq:PFE}
\end{align}

We highlight the strong connection to \Eq{eq:f_dyn_luca}--\Eq{eq:PFE_luca_correct} in Monte Carlo classifier chains. However, in the classification context, the posterior pmf $P$ can be sampled from easily; whereas here, since that might not be the case, we use the auxiliary function $f$ to propose samples at each step in the chain.



There are important differences from typical applications of particle filters, namely 1) in our case the model (\Eq{eq:model}) is learned from the data (i.e., no domain knowledge assumptions), 2) a single observation $\x$ is relevant to an entire sequence of states $y_1,\ldots,y_L$ (that is to say, the isotemporal case), and 3) the full cascade is considered rather than the standard single-order Markovian model.



\subsection{Training}


There are two functions which we need to learn; as according to \Eq{eq:model}: the transition function $f$ and observation function $\ell$. Unlike the vanilla particle filter model described above, we wish to take into account the chain \emph{cascade}, 
\[
	f(y_j|y_1,\ldots,y_{j-1})
\]

Any suitable method for modeling this density can be considered, as long as we are able to sample from it (the methods given in subsections \ref{ssec:1}--\ref{sec:DE} are applicable here). Except, in this case we may remove $\x$ which simplifies and speeds-up sampling. 

For function $\ell$ we do only need to evaluate the function up to some normalizing constant, and thus we have more flexibility in choosing a function to learn. Since we involve observation $\x_i$, predictive power is particularly important at this step. Any method giving accurate 
\[
	\ell(\x) \propto p(y_j|\x,y^{(m)}_1,\ldots,y^{(m)}_{j-1}) 
\]
may be considered. 

There is no need to impose a particular choice at this stage -- we will address options in the experimental investigations in \Sec{sec:experiments}. 

\subsection{Inference}

\Code{code:MSMC} elaborates our Sequential Monte Carlo (SMC) scheme. It can be specifically designed to measure posterior $\p(\y|\x)$; namely to approximate the complex integrals involving of type \Eq{eq:expectation}  
which represents a MMSE estimator, or other loss functions that we are more interested in such as the mode. 

Note that the Effective Sample Size (ESS) approximations is used to decide when the resampling step (and the MCMC/AIS schemes if required) as thus prevent sample degeneration (i.e., error propagation), either 
\[
	\widehat{ESS}({\bar w}_{1:M})=\frac{1}{\sum_{i=1}^M {\bar w}_{i}^2} \quad\text{or}\quad \widehat{ESS}({\bar w}_{1:M})=\frac{1}{\max {\bar w}_{i}}
\]
is typically appropriate \cite{ESSluca}. 

Step \ref{StepMCMC_AIS} in the algorithm is not strictly necessary, nevertheless this step may be useful in cases where the sequential scheme is struggling, and we include a detailed description in \ref{sec:appendix}, of the application of $N$ independent MCMC schemes, specifically, $N$ Metropolis-Hastings (MH) methods (at the $j$-th iteration of the SMC algorithm above).  

Of course, if a model efficiently meets \emph{both} constraints for $f$ \emph{and} $\ell$ (as just mentioned above) we can use the same model for both, thus simplifying \Eq{eq:trans_weights} and recovering the approach described in \Sec{sec:smcrc}.
\begin{algorithm}
	\caption{	\label{code:MSMC} PFC: Sequential Monte Carlo (SMC) Method for PRCs}
	\begin{itemize}
		\item INPUT: 
			\begin{itemize}
				\item $\p_j$, $f_j$, for all $j$, from training stage
				\item  $w^{(m)}_0=\frac{1}{M}$ for all $m$.  
				\item $\eta \in [0,1]$ for ESS approximation 
			\end{itemize}
		\item For $j=1,\ldots,L$:
			\begin{enumerate}
				\item  For $m=1,\ldots,M$:
					\begin{itemize}
						\item Draw samples 
							\(
							y_j^{(m)}\sim f(y_j| \yt_1^{(m)},\ldots,\yt_{j-1}^{(m)})
							\)
						\item Compute the transition weights 
							\begin{equation}
								\label{eq:trans_weights}
								w^{(m)}_{j}=w^{(m)}_{j-1} \frac{\ell(y_j^{(m)}|\x,\yt_{1}^{(m)},\ldots,\yt_{j-1}^{(m)})}{f_j(y_{j}^{(m)}| \yt_{1}^{(m)},\ldots,\yt_{j-1}^{(m)})}
							\end{equation}
					\end{itemize}
				\item\label{StepMCMC_AIS} If $\widehat{ESS}({\bar w}^{(1:M)}_{j})\leq \eta M$: 
					\begin{enumerate}
						\item Resample $\{\yt_{j}^{(1)},\ldots,\yt_{j}^{(M)}\} \sim \{y_j^{(1)},\ldots,y_j^{(M)}\}$ according to normalized weights
							\label{ResStep} 
							\(
							{\bar w}^{(m)}_{j}= \frac{1}{\Zest} w^{(m)}_{j}
							\) where $\hat Z = \sum_{i=1}^M w^{(i)}_{j}$ (i.e., $y_j^{(m)}$ is resampled with probability $\bar w^{(m)}_{j}$; with replacement).
						\item  Set $w^{(m)}_j \gets \frac{1}{M}\Zest$ for all $m$ as per \cite{GIS18}.
						\item\label{StepMCMC_AIS} (Optional) Apply $K$ steps of an MCMC or AIS method (\ref{sec:appendix}); and set (after $K$ iterations)
							\[
								\{\yt_{j}^{(1)},\ldots,\yt_{j}^{(M)}\} \gets \{\yt_{j,K}^{(1)},\ldots,\yt_{j,K}^{(M)}\}
							\]
					\end{enumerate}
			\end{enumerate}
		\item OUTPUT: Prediction $g\left(\{\y^{(m)}\}, \{\w^{(m)}\} \right)$ as per \Eq{eq:the_general}, \\ where $\ypred^{(m)} = \{\yt_{1}^{(m)},\ldots,\yt_{L}^{(m)}\}$
	\end{itemize}
\end{algorithm}







\begin{figure}
    \begin{subfigure}[b]{0.5\textwidth}
		\includegraphics[width=1\textwidth]{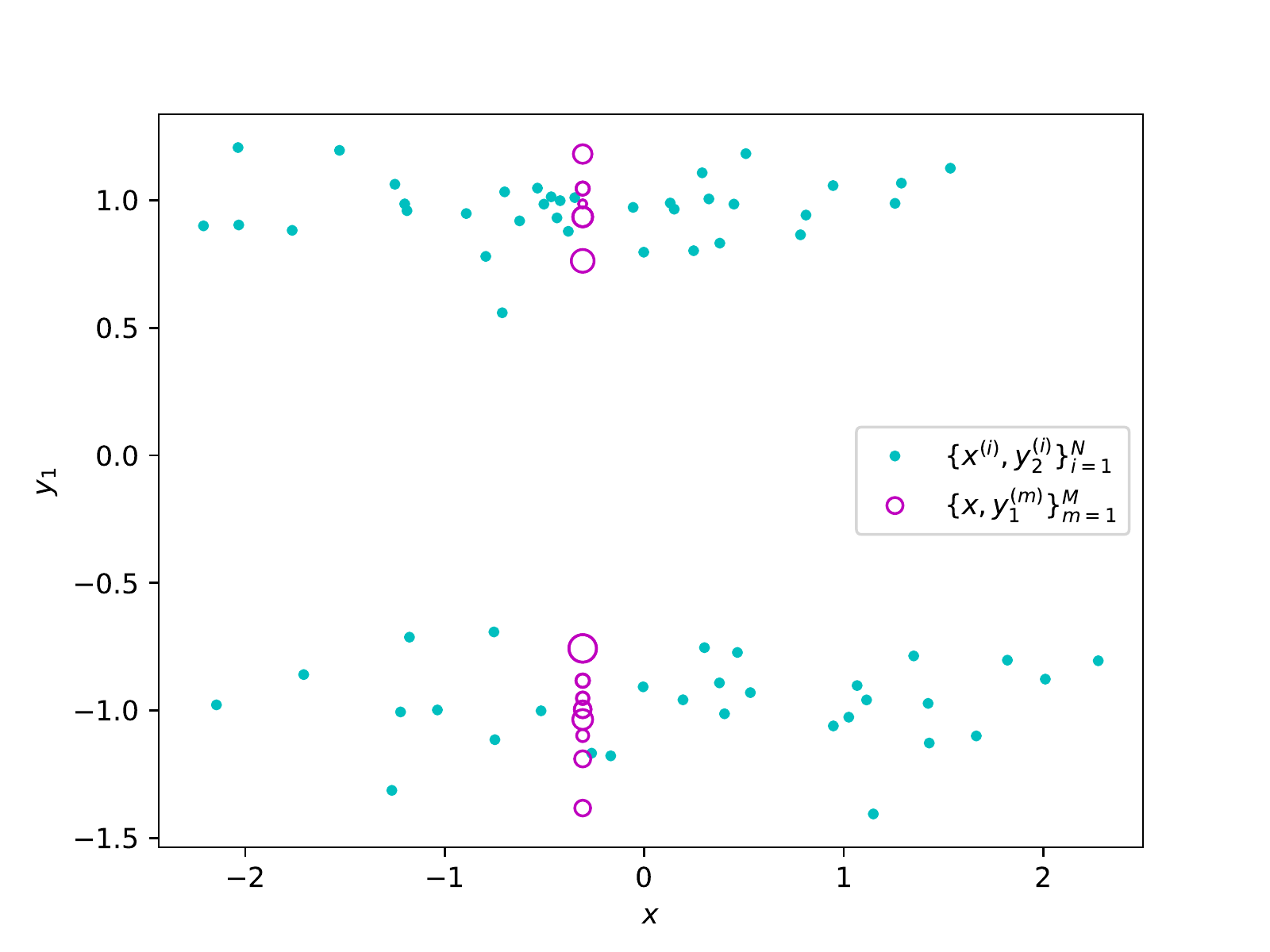}
		\caption{$y_1 \sim f_1(\cdot|x)$}
        \label{fig:m.a}
    \end{subfigure}
    \begin{subfigure}[b]{0.5\textwidth}
		\includegraphics[width=1\textwidth]{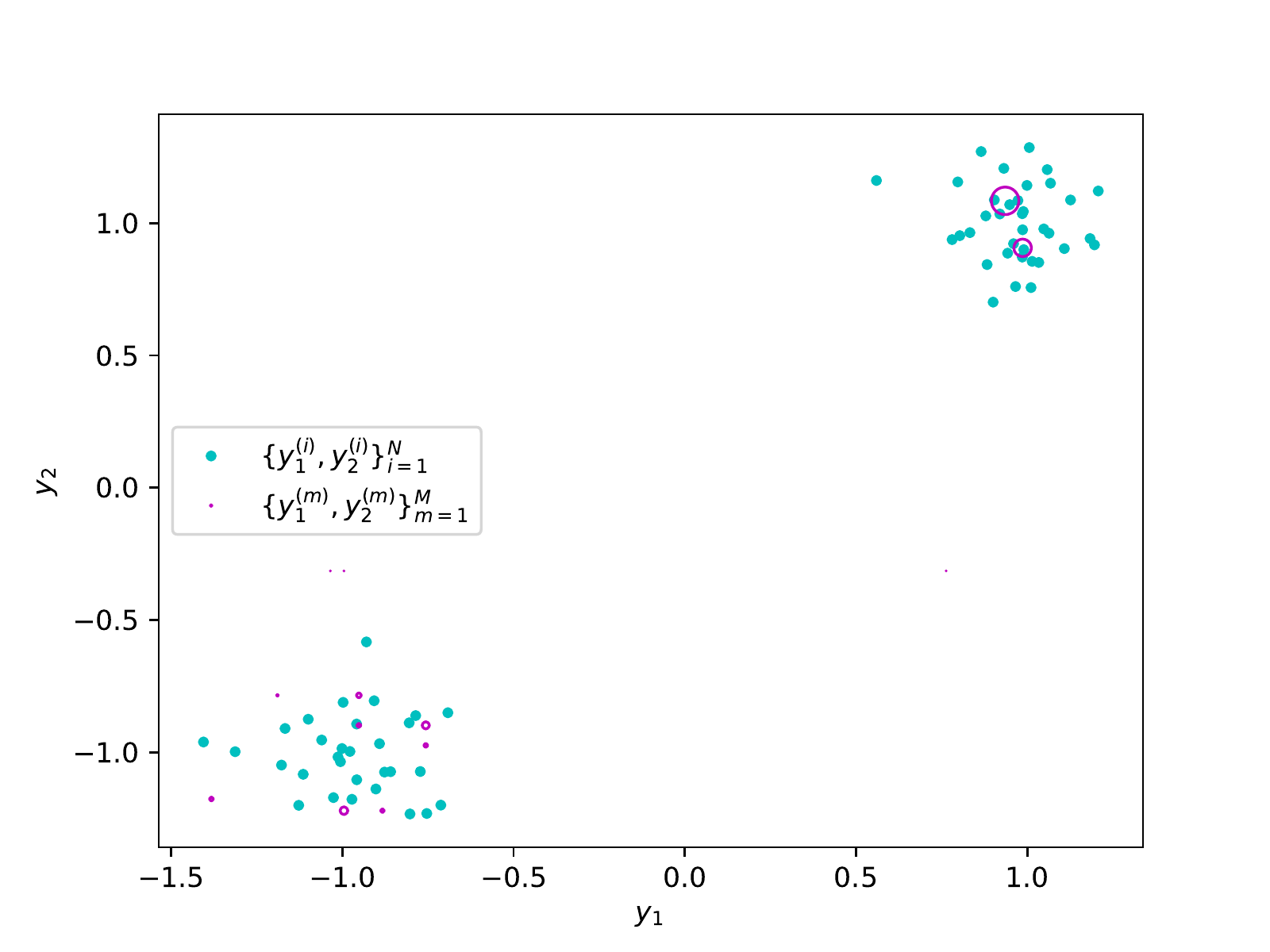}
		\caption{$y_2 \sim f_2(\cdot|x,y_1)$}
        \label{fig:m.b}
    \end{subfigure} \\
    \begin{subfigure}[t]{0.5\textwidth}
		\includegraphics[width=1\textwidth]{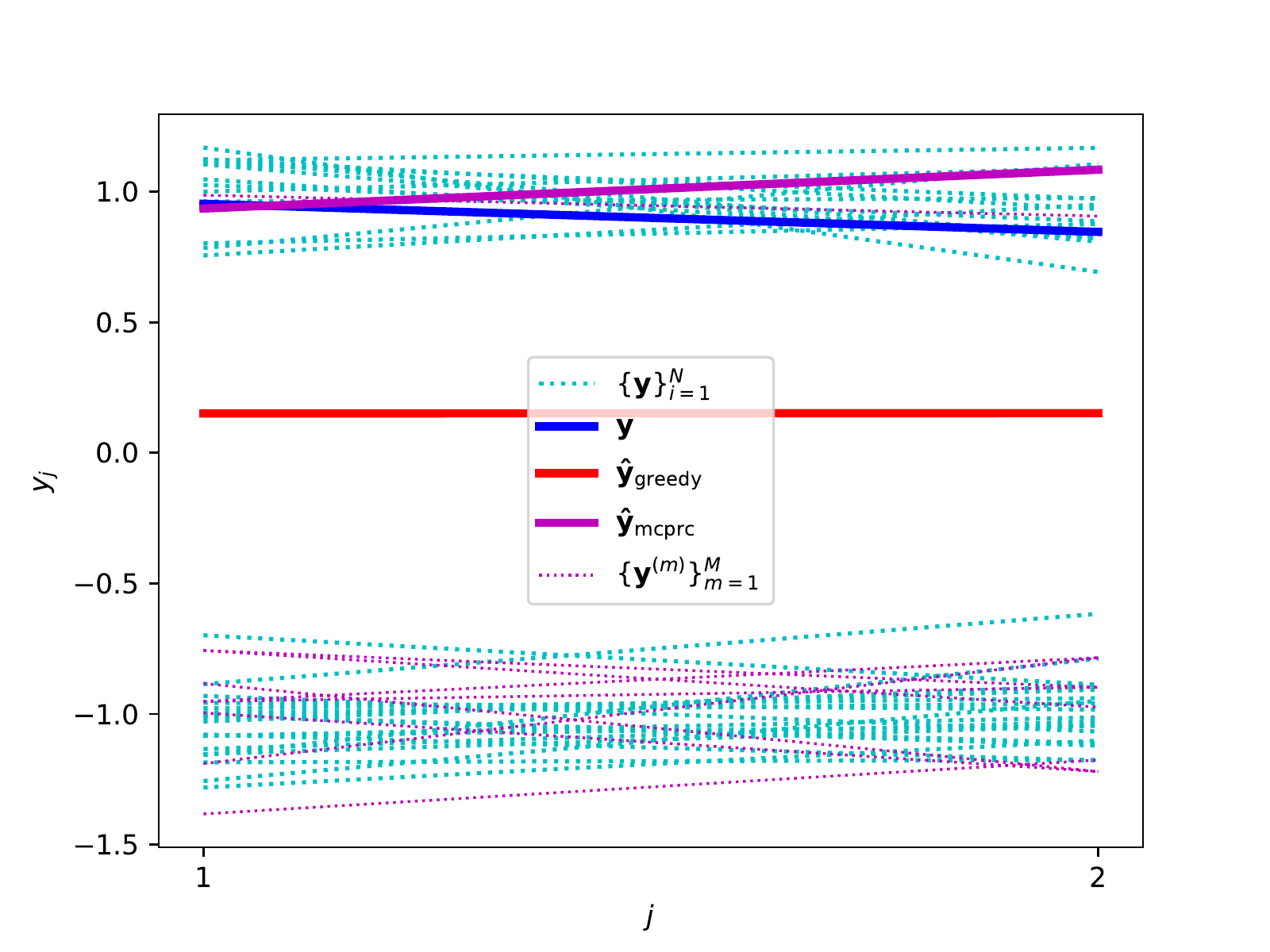}
		\caption{$\ypred = [\yp_1,\yp_2]$}
        \label{fig:m.c}
    \end{subfigure}
    \begin{subfigure}[t]{0.5\textwidth}
		\includegraphics[width=1\textwidth,height=0.75\textwidth]{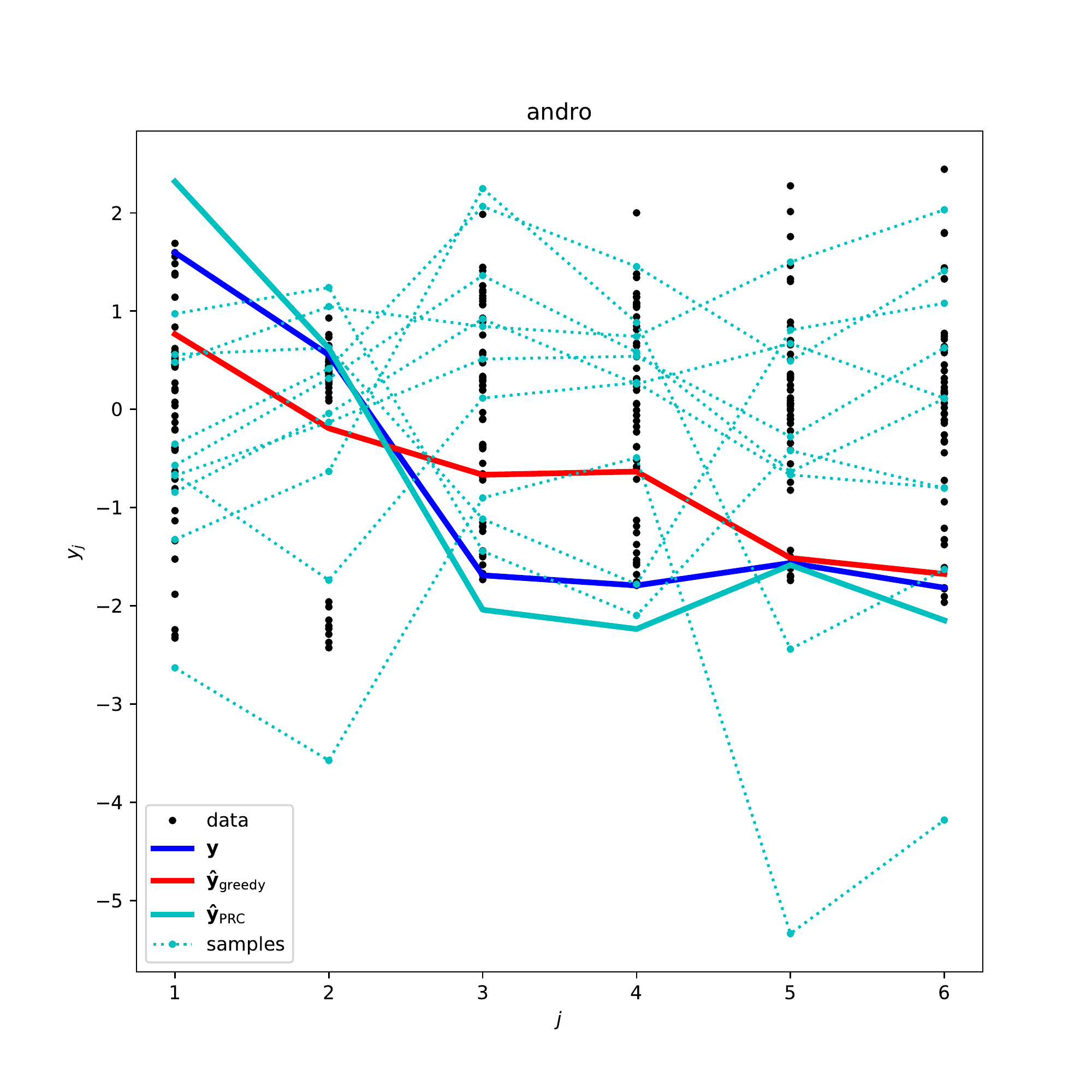}
		\caption{$\ypred = [\yp_1,\yp_2]$}
        \label{fig:m.d}
    \end{subfigure}
	\caption{\label{fig:main_figure} The first three figures above related to the synthetic data shown in \Fig{fig:demo2} (and the forth the Andro dataset). For a given test instance $x$, samples (shown in {\color{magenta}magenta}) are taken across the chain (subfigs.~(\fref{fig:m.a}) and (\fref{fig:m.b}) for $f_1$ and $f_2$ respectively) using function $f_j$ which is learned from the training data (shown in {\color{cyan}cyan}). A separate function evaluates the fitness of these samples, providing weight $w^{(m)}$, as shown implicitly in the size label of each sample. Samples can be viewed as trajectories (\fref{fig:m.c}, and -- for the Andro dataset -- \fref{fig:m.d}), and from this a final trajectory is decided as a prediction. Notice that in this dataset the true trajectory (denoted in {\color{blue}blue}) can be approximated by our method, whereas this is never the case under greedy chains (at least as clearly shown in the synthetic dataset). We remark also the bimodal nature of the distribution of Synth which is difficult to capture with standard regression methods. Details in \Code{code:MSMC} and \Sec{sec:experiments} in general.}
\end{figure}

\section{Experiments}
\label{sec:experiments}

In this section we compare some of the approaches we discussed, identified, and developed above. Namely, we compare independent regression models (IR), regressor chains with greedy inference (RC) with our Monte Carlo methods, discussed in \Sec{sec:PRC} as well as further developed as a Particle Filter  in \Sec{sec:smcrc} (PFC). We compare different base estimators. Recall that PFC takes both a model for sampling and a model for evaluation -- not necessarily from the same model class. These models are summarized and denoted as follows:
\begin{center}
	\small
		\begin{tabular}{ll}
			\hline
			Key & Algorithm \\
			\hline
			    IR		& Independent Regression \\
				RC		& Greedy Regressor Chains \\
				MC		& Probabilistic Regressor Chains (Monte Carlo) \\
				PF		& Probabilistic Regressor Chains (Particle Filter) \\
			\hline
				B      & Bayesian Regression \\
				K      & Kernel Ridge Regression (Gaussian Kernel) \\
				       & grid search $\alpha \in \{1,0.1,0.01,0.001\}$ and $\gamma \in \{0.01,0.1,1,10,100\}$ \\
				N & Discretized label space, 30 bins (classes per label) \\
				  & \ldots with Neural Network Classifier (2 hidden layers each of size 30) \\
				R & \ldots with Random Forest Classifier, 100 estimators \\
			\hline
		\end{tabular}
\end{center}
such that (in the tables of results, \Tab{tab:results}), IR.B denotes independent Bayesian regression, PF.R/B denotes particle filter chains with a discretized random forest base classifier for sampling, and Bayesian regression for evaluation; and so on. Note that MC and PF are \emph{always} set to maximize the $0/1$ approximation (by selecting a mode), whereas IR and RC (by default) maximize MSE, i.e., predict the mean. In both cases we consider $M=100$ samples/particles per test example; $\eta=0.1$.

IR and RC are implemented in the well-known Scikit-Learn framework\footnote{\url{https://scikit-learn.org}}. We implemented our novel contributions using the Scikit-Multiflow framework\footnote{\url{https://scikit-multiflow.github.io/}} \cite{MultiFlow}; which is based on Scikit-Learn. If not explicitly stated, then default parameters are used. 

\begin{figure}[h]
	\centering
	\includegraphics[width=0.45\textwidth]{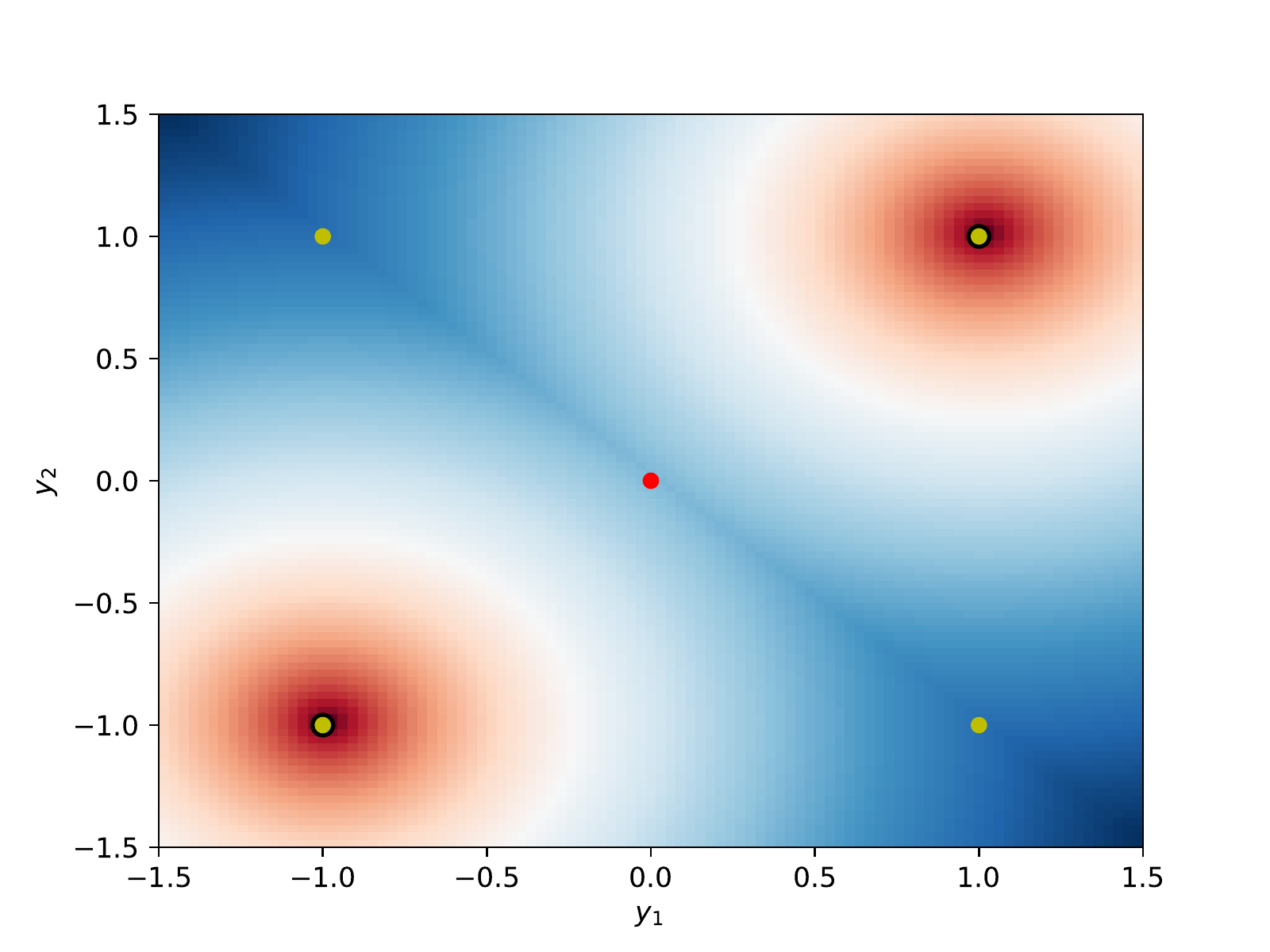} 
	\includegraphics[width=0.45\textwidth]{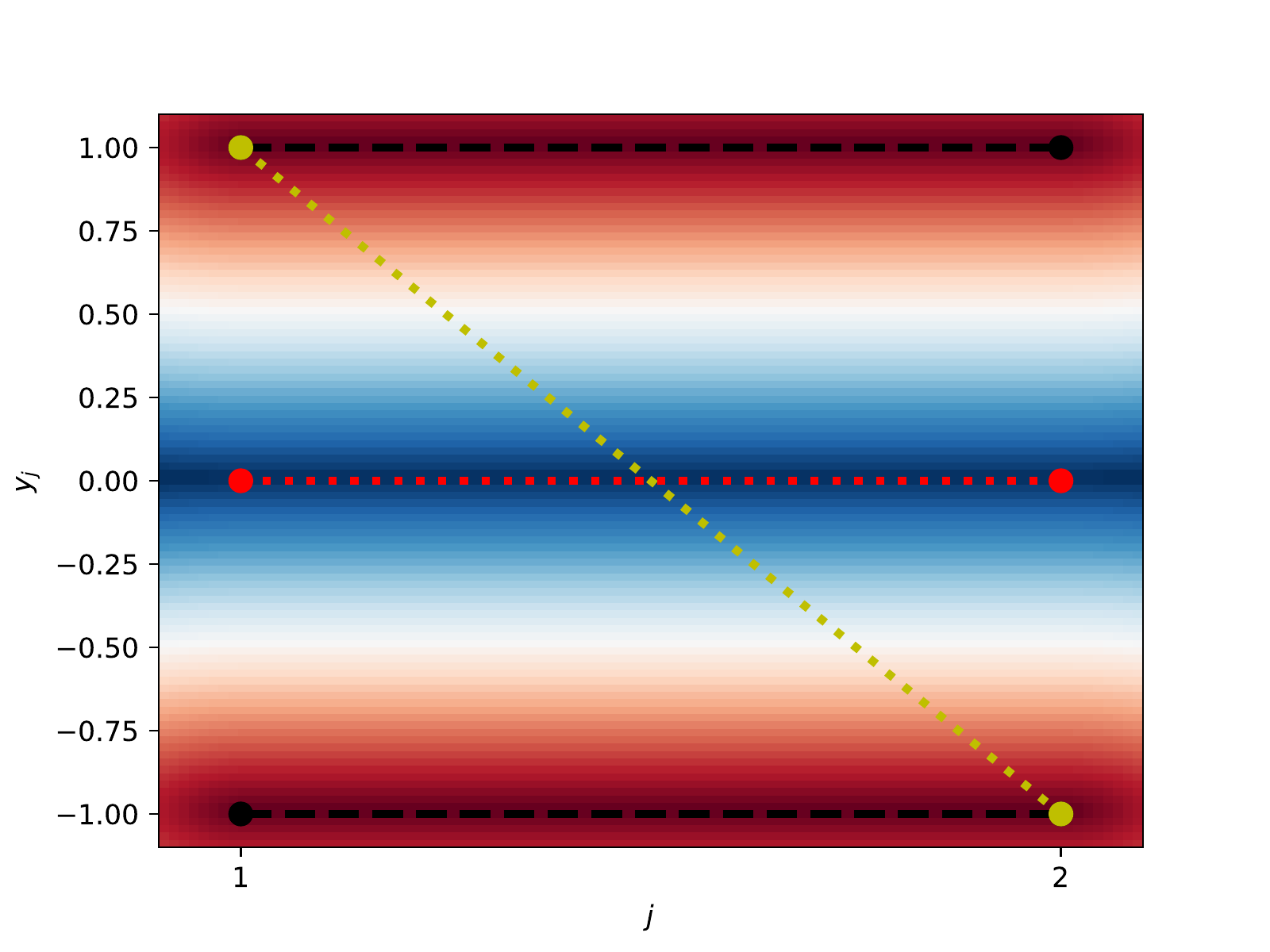} 
	\caption{\label{fig:demo2} A bimodal joint distribution over two labels $L=2$. We suppose that both modes are equally probable given $x$. Left: The true density. Right: hypothetical ``paths'' across $j=1,2$. The black points/lines show estimates under a MAP estimator, red from an estimator of MSE, and yellow one possible results under MAE. }
\end{figure}

We carry out an empirical evaluation on synthetic and real-world datasets. 
The real-world sets are described and referenced in \cite{MTRexpansion}\footnote{Available online: \url{http://mulan.sourceforge.net/datasets-mtr.html}}; covering a number of real-world applications involving predicting the multi-components of sea-water, residential buildings, concrete pouring, and natural resources; so as to over sufficient variety. The synthetic dataset is described and shown in \Fig{fig:demo2}. 

We consider the following loss metrics: 
\[
	\textsf{MSE} = \frac{1}{N} \sum_{i=1}^N \sum_{j=1}^L (y_{ij} - \yp_{ij})^2  \text{\quad and \quad} \textsf{MAE} = \frac{1}{N} \sum_{i=1}^N \sum_{j=1}^L |y_{ij} - \yp_{ij}|
\]
(mean squared error and absolute error, respectively) and denoting
\[
	\textsf{$0/1$} \approx \frac{1}{N} \sum_{i=1}^N \CaseOtherwise{0}{\|\ypred_i - \y_i\|_2 < c}{1}
\]
as an approximation of mean $0/1$ loss (inverse accuracy) which approaches the $0/1$-loss estimate as constant $c$ goes to $0$. This loss is designed to reward models which find a joint mode, which should correspond to a path which is likely to occur in practice (e.g., in the training data). 


Averaged results over 10-fold cross validation are provided in \Tab{tab:results} for the different metrics. 

%

\begin{table}
	\centering
	\caption{\label{tab:results}Results of 10-fold cross validation. 
	}
	\small
	\begin{center}
		MSE:
	\end{center}
	\begin{tabular}{llllllllll}
\hline 
Dataset & $L$ & IR.B & IR.K & RC.B & RC.K & MC.B & MC.D & PF.R/B & PF.N/B \\ 
\hline 
                Synth &   2 & 1.03 & 1.02 & 1.04 & 1.05 & 1.44 & \textbf{1.02} & 1.51 & 1.42 \\ 
                Andro &   6 & 0.60 & 0.24 & 0.50 & \textbf{0.21} & 0.77 & 0.91 & 1.04 & 1.02 \\ 
                  EDM &   2 & 0.61 & 0.43 & 0.60 & \textbf{0.42} & 0.97 & 0.80 & 0.98 & 1.17 \\ 
                  ENB &   2 & 0.10 & 0.01 & 0.10 & \textbf{0.01} & 0.15 & 0.58 & 0.16 & 0.20 \\ 
                 Jura &   3 & 0.38 & \textbf{0.35} & 0.38 & 0.35 & 0.57 & 0.77 & 0.58 & 0.50 \\ 
                Slump &   3 & 0.49 & \textbf{0.40} & 0.49 & 0.40 & 0.80 & 0.85 & 0.64 & 0.54 \\ 
\hline 
\hline 
Avg Rank &   & 3.67 & 2.67 & 2.33 & 1.67 & 6.00 & 6.33 & 6.67 & 6.67 \\ 
\hline 
\hline 
\end{tabular} 

	\begin{center}
		MAE:
	\end{center}
	\begin{tabular}{llllllllll}
\hline 
Dataset & $L$ & IR.B & IR.K & RC.B & RC.K & MC.B & MC.D & PF.R/B & PF.N/B \\ 
\hline 
                Synth &   2 & 1.01 & 1.00 & 1.01 & 1.01 & 1.00 & 1.00 & 0.91 & \textbf{0.89} \\ 
                Andro &   6 & 0.62 & 0.33 & 0.56 & \textbf{0.31} & 0.70 & 0.76 & 0.95 & 0.75 \\ 
                  EDM &   2 & 0.63 & 0.46 & 0.62 & \textbf{0.44} & 0.80 & 0.62 & 0.88 & 0.87 \\ 
                  ENB &   2 & 0.22 & 0.07 & 0.22 & \textbf{0.06} & 0.28 & 0.64 & 0.32 & 0.36 \\ 
                 Jura &   3 & 0.41 & \textbf{0.39} & 0.41 & 0.39 & 0.56 & 0.66 & 0.53 & 0.51 \\ 
                Slump &   3 & 0.54 & \textbf{0.42} & 0.55 & 0.43 & 0.67 & 0.76 & 0.61 & 0.56 \\ 
\hline 
\hline 
Avg Rank &   & 4.00 & 3.50 & 3.00 & 2.50 & 5.33 & 5.83 & 6.00 & 5.83 \\ 
\hline 
\hline 
\end{tabular} 

	\begin{center}
		$\approx 0/1$ ($c=0.1$):
	\end{center}
	\begin{tabular}{llllllllll}
\hline 
Dataset & $L$ & IR.B & IR.K & RC.B & RC.K & MC.B & MC.D & PF.R/B & PF.N/B \\ 
\hline 
                Synth &   2 & 1.00 & 1.00 & 1.00 & 1.00 & 0.94 & 1.00 & \textbf{0.51} & 0.59 \\ 
                Andro &   6 & 0.98 & 0.72 & 0.98 & \textbf{0.69} & 1.00 & 1.00 & 1.00 & 0.98 \\ 
                  EDM &   2 & 0.87 & 0.69 & 0.86 & \textbf{0.64} & 0.88 & 0.78 & 0.82 & 0.70 \\ 
                  ENB &   2 & 0.21 & 0.02 & 0.21 & \textbf{0.01} & 0.32 & 0.81 & 0.39 & 0.47 \\ 
                 Jura &   3 & 0.73 & \textbf{0.70} & 0.73 & 0.70 & 0.88 & 0.98 & 0.89 & 0.87 \\ 
                Slump &   3 & 0.82 & \textbf{0.65} & 0.82 & \textbf{0.65} & 0.91 & 0.91 & 0.76 & 0.81 \\ 
\hline 
\hline 
Avg Rank &   & 3.83 & 3.67 & 3.83 & 4.00 & 5.17 & 4.33 & 5.17 & 6.00 \\ 
\hline 
\hline 
\end{tabular} 

\end{table}

%

\section{Discussion}
\label{sec:discussion}

In this section we discuss empirical results. We look at these results with the goal of drawing conclusions about the behaviour of regressor chains in general, and secondly as justifying both acceptable performance and usability of our proposed methodologies. 


The empirical results confirm that greedy regressor chains (RC) shows little to no advantage against independent estimators when a linear base model is used (only a small exception under the Andro dataset, if we are to compare RC.B and IR.B). These findings are completely in line with the analysis so far: classification models involve an inherent non-linearity (such as for example the sigmoid function in logistic regression) which adds predictive power via the chain structure, but this not inherently the case in regression. 


We do not need to discuss the power of non-linear modeling for regression, as this is an elementary concept, but it is particularly interesting to observe how well regressor chains with a non-linear base learner (e.g., RC.K) can perform better on average (i.e., in terms of average rank) against its independent counterpart (IR.K). It is in this sense that we begin to find an argument to use regressor chains. 


Although regressor chains may be effective, we point out the risk of degeneration of estimates across the chain with poorly regularized base models. Indeed, we found this using standard stochastic gradient descent linear regression, estimates diverged so far (obtaining greater than 1000 MSE) that it there was no point to include them in the table. This effect is well-mentioned in many analyses of classifier chains in the literature under the term of \emph{error propagation}, but in a multi-output binary classification setting, the posterior $P(y_j|\x,y_{j-1},\ldots,y_1)$ such propagation is always constrained between $0$ and $1$. However, under RCs the trajectory may become increasingly lost and isolated (from the true path) in $\R^L$ space as prediction progresses along the chain. Our choice of Bayesian regression adds some regularization which counters this. Nevertheless it is also one of the issues we took into account with our developments of probabilistic regressor chains. 



Results show that acceptable predictive performance can be usually obtained by Monte-Carlo approaches (MC, PF), although on average it does not achieve top results overall. At first glance it seems difficult to justify either of these two methods in either of their configurations, but, we can take a closer look at particular evaluation metrics. We find that when it is important to find modes of the posterior ($0/1$), MC/B and PF./B outperform RC.B about half of the time. We can speculate that this would be similar in a hypothetical comparison of MC./K but the implementation we used (from ScikitLearn) does not offer sampling from this method. Furthermore, such a kernel-based approach has its own disadvantages (quadratic complexity, difficult application to incremental learning, and so on) which are limiting in modern settings of large and dynamic datasets. In addition, we can emphasise the important result on Synth: 1.00 (RC.K) vs 0.51 (PF.R/B) is one of the largest differences in performances obtained (actually 0.50 would be the Bayes optimal result for this data). Finally, and perhaps most importantly: Greedy RC provides almost no form of interpretation. 




Results are on the Synth dataset are worth exploring in more detail (shown in \Fig{fig:demo2}; results shown in the first row in \Tab{tab:results}; detailed step-by-step illustration of performance in \Fig{fig:main_figure}). In particular, notice that any path crossing a low density region is unlikely to exist in practice, even though it provides the best estimate under MSE. In any kind of data with a multi-modal density in the output space, the our developed PF approach is highly suited, as it is able to track the path-evolution across different modes along its pass of the chain. Furthermore, it is able to offer interpretation or \emph{explainability} of results by showing actual paths taken, and the density estimate via the cloud of particles; see, e.g., \Fig{fig:main_figure}. 

It is a pity that the benchmark datasets we used as per related studies, do not seem complex enough to reveal this behaviour, although it is easy to provide examples of real-world possibilities. For example, in the prediction of trajectories for vehicles, it is useful only to estimate paths which do not cross out of transportation axes (for example, a truck does not pass through a river but over one of the bridges that cross it; for example as covered in \cite{TSPF} in the discrete-waypoint scenario). Furthermore, it can be interesting to produce a set of possible trajectories rather than a single estimation.  
Likewise, in time series forecasting and anomaly detection, we may want to consider different hypotheses, rather than outputting a single estimator of maximum likelihood. 

The need for interpretable models is certainly of increasing interest, as machine learning methods are used in more sectors, for example medical and security, where it is often essential to be able to provide detailed explanation of results. 

For simplicity, in the Synth data, the observation $x$ does not affect the prediction for $y_2$. It means that if that variable was added (in a real-world case) following building and deployment of the model predicting $y_1$, it would be more efficient to update the multi-output with a chain than an independent classifier. Of course on this dataset any discussion of efficiency is not warranted, but as the number of labels and instances grows, we can see chain methods operating as a kind of transfer learning, adapting partially pre-trained models by adding links from the outputs of those models. A further discussion is beyond the scope of this work.

\section{Related Methods and Additional Considerations}
\label{sec:related}

To the best of our knowledge this is the first work treating regression chains in depth, and particularly from a probabilistic point of view with a sequential Monte Carlo approach. Nevertheless, since it is tackling a well-established problem (i.e., multi-output regression) it is equally important to contribute a discussion of related methods in other areas, and additional considerations treated in the classifier-chains literature already, as we do in this section. 

\subsection{Neural networks}
\label{ref:nn}

A residual neural network (ResNet, \cite{ResNet}) includes skip layers similar to the way a chain model does, except layer-wise rather than node-wise. It has obtained noteworthy performance on deep learning tasks. The original paper (2015) obtained good performance in image classification, though has also indeed been studied in the context of regression (such as in \cite{DeepRegression}). Aside from its prevalence in classification, another difference is that in ResNets only the last layer is used for prediction, with other labels simply forming internal nodes. Although, similar multi-label architectures (including, e.g., \cite{ADIOS,DCC2}) have been used which consider each (or at least several nodes/layers) as the targets, i.e., multiple targets. Although this work deals primarily with classification problems, this deep-learning approach tackles well one of the issues faced by regressor chains: needing strong non-linearities to serve as useful representations, as highlighted in our discussion and results. Other neural models have been considered specific to the multi-output regression case, including ensemble settings, e.g., \cite{MultiTargetRegressionNeural}.

\subsection{Probabilistic models}

A closely related approach to the inference in probabilistic regressor chains can be seen as performing a standard regression with noisy inputs \cite{Dellaportas95,Johnson2018}. Indeed, suppose that $L=2$, and
\begin{eqnarray}
y_1&\sim& p_1(y_1|\x), \nonumber \\
y_2 &\sim& p_2(y_2|y_1,\x).
\end{eqnarray}
Under the assumption of additive noise, we can write 
 \begin{eqnarray}
y_1&=& h_1(\x)+ \epsilon_1 \\
y_2 &=&  h_2(\x,y_1)+\epsilon_2,
\end{eqnarray}
where $h_j$ are the regression models. For performing a proper inference, all the statistical features of $y_1$ should be taken into account (i.e., the uncertainty), hence a regression problem with noisy inputs. Gaussian Processes (GPs) provide a method relevant to this context, e.g., \cite{Quin03,DALLAIRE20111945}, and in particular the idea of warped GPs \cite{WarpedGP,NIPS2012_4494,DamianouDP13} (a kind of hierarchical GP, thereby giving a kind of `depth'). Most of these cases are elaborated where $L=2$. We remark (also with regard to discussion above) that there is of course no clear separating definition between (deep) neural network and probabilistic models. 

\subsection{State space models}

The previous considerations have direct application to the inference and prediction in so-called state-space models. Such models are formed by a transition dynamic equation and an observation equation\footnote{Once more, we have opted for notation where $x$ is the observation}, 
\begin{gather}
\left\{
\begin{split}
	y_{t+1} &= h_d(y_t)+\epsilon_{d,t},  \\
	x_{t+1} &= h_o(y_{t+1})+\epsilon_{o,t}, 
\end{split}
\right. 
\end{gather}
where $y_t$ is the \emph{state} at time $t$, and $\epsilon$ is perturbation noise. Note that even if the mappings $h$ and noises $\epsilon$ are known, the inference and prediction in this stochastic system can be interpreted, at each $t$, as a noisy input regression problem since
\begin{equation}
x_{t+1} = h_o(h_d(y_t)+\epsilon_{d,t})+\epsilon_{o,t}, 
\end{equation}
showing strong parallels with the noisy regression model above. Furthermore, when the elements of the dynamic system are not deterministically known the problem becomes even more complex. In \cite{Deisenroth09,Bijl2016SystemIT}, authors suggest to model $h_d$, $h_o$ as two GPs. 

\subsection{Chain order} 

A natural question which arises is -- does the order of the chain affect results? This question has been answered in the classification context (in fact, it \emph{does} affect results) in various other works, mainly by the use of random ensemble methods (e.g., \cite{ECC2,RAkEL2}; each model with a different/random order), hill-climbing approaches (e.g., \cite{CCAnalysis,MCC2}; the best of a number of trialed orders is taken) and methods based on label dependence (e.g., \cite{AstarCC,UPM} and references therein). Most of these methods are also directly applicable to the regression case; indeed, the chain-order space and complexity is exactly the same, and thus ensemble and hill-climbing methods need no specific adaptation; label-dependence based methods only need to consider dependence among continuous variables rather than discrete. Thus, there is no reason to suggest why the regression context needs special treatment, and for this reason we do not specifically readdress this consideration in this work.



\subsection{Multiple passes} 

Another aspect to consider is why not pass over the set of variables (i.e., through the chain) multiple times per test instance. This is an interesting proposal and would be an easy extension to our methods: simply iterate over the chain a second time, and plug in training labels $y_{L+1} = y_1,\ldots, y_{L+L} = y_L$ and treat them thenceforth as the target labels of interest. We notice that one can view this as a special case of \textit{regressor stacking} as described in \cite{HanenMORSurvey} (in the general case, we simply feed all predictions as inputs to a second model --  not necessarily a chain model). On the other hand, time complexity increases significantly with each pass along the chain. One can also see a connection to recurrent neural networks (RNN), as noticed in recent work by \cite{RNNMLC}, unrolled across time. 

From the probabilistic perspective, an approach of many passes can be seen as a kind of Gibbs sampling (where the graph is undirected). In fact, an undirected and fully connected network (rather than a `chain') removes the question of label order entirely. This has been developed in the context of \textit{conditional dependency networks} as presented by \cite{CDN} (for multi-label classification). We notice that this framework is also applicable to the regression case, whenever sampling from the conditional is possible; namely, (as in \cite{CDN}) using Gibbs sampling to estimate the marginal mode: 
\begin{align}
	y^{(m)}_j     & \sim p(y_j|\x,y_{\neq j}) \label{eq:f_dyn_gibbs} \\
	p(y_j)      & \approx \frac{1}{M - M_0} \sum^M_{m=M_0} y^{(m)}_j \label{eq:PFE_gibbs}
\end{align}
(having adjusted the equations for continuous target variables) where $M_0$ indicates the initial samples during the burn-in time which are not considered. 
This is relatively straightforward to incorporate in the regression setting. It is, however, worth emphasising that 
(as found empirically in \cite{MCC2}, for example), that the number of samples must be relatively much greater since a burn-in time of $M_0 > L$ samples is required, and $M \gg L$. 

\subsection{Other areas of application}

Regressor chains can be applied to any problem involving multiple continuous output variables. Time series forecasting is a natural application, e.g., \cite{Quin03}. Previously classifier chains was applied to a discretized version of route prediction in urban traffic modeling \cite{ChainInTime}; and an application of regressor chains in continuous space would be perhaps even more natural. 

One interesting similarity is to the application of Monte Carlo tree search for continuous action spaces in reinforcement learning (a good example of this is given in \cite{MCTS_CAS}). Forming a tree to search on top of a continuous space is part of the problem we tackle. We notice that the authors of this cited work use a kernel regression approach which is what we have empirically found to work well in this work on regressor chains. Unlike our study, there is no use of a chain in the sense of a cascade, which is the main focus of this work. Clearly investigation of the use of regressor chains in such related areas could be promising.


\section{Conclusions}
\label{sec:conclusion}

In this work we have looked at extending the approach of chaining models to the regression context. This was motivated by the fact that, although the idea had attracted initial interest and found potential applicability to multi-output regression tasks, the behaviour of regressor chains was found not to emulate that of classifier chains, thus warranting this in-depth study, to identify, unravel, and overcome the weaknesses of this application. 

We identified, discussed, and dealt with several important issues in this regard: 
\begin{itemize}
	\item Regressor chains are only useful when non-linearities are used in the base models, and this is equally applicable to the standard multi-output case, and the case where attribute values arrive in sequence 
	\item It is difficult to justify regressor chains for MSE estimation, unless paying particular attention to the previous point. 
	\item Error propagation (progressive divergence from the true path estimate) can be much more severe in the regression case due to the unbounded output space 
	\item Greedy inference in regressor chains, on top of the above points, is unable to provide useful representation or interpretation about the output space. 
\end{itemize}



To our knowledge, this is the first work which looks in-depth at regressor chains in the probabilistic context. We surveyed a number of applicable methods, and developed our own based on sequential Monte Carlo methodology, particularly crafted to tackle these identified issues. We use non-linear base models as particular to the `chains' methodology where base models themselves can be considered as a flexible hyperparameter, guided by aspects of the problem domain, rather than a hardwired design -- this, we argue, should not be seen as a `nuisance' parameter but as an attractive feature for adaptation across different areas. We provide a mode-seeking mechanism, rather than only (but as well as, potentially) an estimator for minimum squared error. Error propagation is controlled by a resampling scheme. And interpretation is provided by a tree generated by the point cloud of sample paths, which offer also a description of the underlying density. 

Probabilistic regressor chains have peculiarities that distinguish it from other models, such as the full cascade involving all outputs and hyper-parametrization of the base models. However, to properly place regressor chains in context with the wider literature, we also identified and discussed connections to a range of related work which had not been noticed. 

We may argue that our analysis not only facilitates understanding the performance of regressor chains that we develop, but also earlier throws more light into the performance considerations of classifier chains, modifications of which are still under active development and recent publication. 


Our Monte Carlo methods suggest to be promising especially for tasks where path explainability (i.e., different hypotheses regrading the path taken through output space) is of more importance than outputting the result of a minimum mean squared error estimator. Such tasks include medical research, anomaly detection, de-noising, and missing-value imputation; providing plenty of application lines along which to develop this work further and built additional connections with related areas of the literature. In future work we also intend to explore areas (such as dynamic chains and recurrent models) that are being developed in parallel in the classification context to approach themes relating to chain order and structure. 


\bibliographystyle{plain}
\bibliography{../multilabel,../jesse,../extra,../luca}

\appendix

\section{Parallel Metropolis-Hastings (MH) chains}
\label{sec:appendix}

For completeness, we elaborate the algorithm based on \cite{BUGALLO201536,7544571}, starting from the equal weighted set of samples obtained at step \ref{ResStep} of \Code{code:MSMC}, $\{\yt_{j,0}^{(1)},\ldots,\yt_{j,0}^{(M)}\}$. After $K$ iterations, we obtain the final set of samples, illustrated as follows for the $m$-th particle:

\begin{itemize}
	\item For $k=1,\ldots,K:$
		\begin{enumerate}
			\item Draw  $z \sim q(y|y_{j,k-1}^{(m)})$.
			\item Setting for simplicity $\pj_j(z)=\pj_j(z|\x,{\widetilde y}_{1:j-1}^{(m)})$, accept the movement $y_{j,k}^{(m)}=z$ with probability 
				\begin{equation}
					\alpha=\min\left[1,\frac{\pj_j(z) q(y_{j,k-1}^{(m)}|z)}{\pj_j(y_{j,k-1}^{(m)}) q(z|y_{j,k-1}^{(m)})}\right]
				\end{equation}
				Otherwise, with probability $1-\alpha$, set $y_{j,k}^{(m)}=y_{j,k-1}^{(m)}$
		\end{enumerate}
\end{itemize}

	Therefore final set of samples for an iteration is $\{{\widetilde y}_{j}^{(1)},\ldots,{\widetilde y}_{j}^{(M)}\}=\{{\widetilde y}_{j,K}^{(1)},\ldots,{\widetilde y}_{j,K}^{(m)}\}$.

\end{document}